\documentclass{article}

\usepackage{arxiv}

\usepackage[utf8]{inputenc} 
\usepackage[T1]{fontenc}    
\usepackage{hyperref}       
\usepackage{url}            
\usepackage{booktabs}       
\usepackage{amsfonts}       
\usepackage{nicefrac}       
\usepackage{microtype}      

\usepackage{graphicx}
\usepackage{doi}
\usepackage{amsmath,amssymb}
\usepackage{algorithmic}
\usepackage{graphicx}
\usepackage{textcomp}
\usepackage{xcolor}
\usepackage{multirow}
\usepackage{caption}
\usepackage{subcaption}

\title{Experience feedback using Representation Learning for Few-Shot Object Detection on Aerial Images\\}

\author{
	\textbf{Pierre Le Jeune}\\
	COSE \& LIPN  \& L2TI\\
	Université Sorbonne Paris Nord\\
	\texttt{pierre.le-jeune@cose.fr}\\
	\and
	\textbf{Mustapha Lebbah}\\
    LIPN\\
    Université Sorbonne Paris Nord\\
    \texttt{mustapha.lebbah@lipn.univ-paris13.fr}\\
    \and
    \textbf{Anissa Mokraoui}\\
    L2TI\\
    Université Sorbonne Paris Nord\\
    \texttt{anissa.mokraoui@univ-paris13.fr}\\
    \and
    \textbf{Hanene Azzag}\\
    LIPN\\
    Université Sorbonne Paris Nord\\
    \texttt{azzag@lipn.univ-paris13.fr}\\
}

\date{}

\hypersetup{
pdftitle={A template for the arxiv style},
pdfsubject={q-bio.NC, q-bio.QM},
pdfauthor={David S.~Hippocampus, Elias D.~Striatum},
pdfkeywords={First keyword, Second keyword, More},
}

\begin{document}

\maketitle

\begin{abstract}
This paper proposes a few-shot method based on Faster R-CNN and representation
learning for object detection in aerial images. The two classification branches
of Faster R-CNN are replaced by prototypical networks for online adaptation to
new classes. These networks produce embeddings vectors for each generated box,
which are then compared with class prototypes. The distance between an embedding
and a prototype determines the corresponding classification score. The resulting
networks are trained in an episodic manner. A new detection task is randomly
sampled at each epoch,  consisting in detecting only a subset of the classes
annotated in the dataset. This training strategy encourages the network to adapt
to new classes as it would at test time. In addition, several ideas are explored
to improve the proposed method such as a hard negative examples mining strategy
and self-supervised clustering for background objects. The performance of our
method is assessed on DOTA, a large-scale remote sensing images dataset. The
experiments conducted provide a broader understanding of the capabilities of
representation learning. It highlights in particular some intrinsic weaknesses
for the few-shot object detection task. Finally, some suggestions and
perspectives are formulated according to these insights. 

\end{abstract}


\section{Introduction}
Object detection is a key problem in computer vision. It consists in finding all
occurrences of objects belonging to a predefined set of classes in an image and
classify them. Its applications range from medical diagnosis to aerial
intelligence through autonomous vehicles. Object detection methods automate
repetitive and time-consuming tasks performed by human operators until now. In
the context of Remote Sensing Images (RSI), detection is used for a wide variety
of tasks such as environmental surveillance, urban planning, crops and flock
monitoring or traffic analysis.

Deep learning and especially convolutional neural networks (CNNs) outperform
previous methods on most computer vision tasks and object detection is no
exception. Plenty of methods have been introduced to address this challenge.
Among them, Faster R-CNN \cite{ren2015faster} and YOLO \cite{redmon2016you} may
be the most well-known and studied. Even though this problem is far from being
solved, detection algorithms perform well when provided with sufficient
annotated data. However, this is often not available in practice and the
creation of large dataset for detection requires both time and expertise
preventing the deployment of such methods for many use cases. Another limitation
to the widespread deployment of detection techniques is the lack of
adaptability. Once fully trained, it is hard to modify a model to adapt to new
objects. This is critical for some applications which need to detect different
objects from one usage to another. Aerial intelligence is an example of such
application: each mission may have its specific objects of interest and
therefore a detection model must be adaptable (literally) on the fly. The
overall objective of this work is to be deployed on vertical aerial images. Yet,
large-scale dataset of such images, annotated for object detection, are rare.
RSI are a convenient alternative and provide an accurate estimation of
performance in deployment.

Few-Shot Learning (FSL) techniques have been introduced to address these issues
and deal with limited data. This has been extensively studied for classification
\cite{snell2017prototypical}, \cite{ravi2016optimization}. Its principle is to
learn general knowledge from a large dataset so that it can generalize
efficiently (i.e. quickly and from limited data) on new classes. There exist
different approaches for this task. Representation learning tries to
learn abstract embeddings of images so that the representation space is
semantically organized in a way that makes classification relatively easy (e.g.
with a linear classifier). Meta-learning based methods learn a model (teacher)
that helps another model (student) to perform well based on a limited amount of
data. This is often done by training both network on multiple low-data tasks
(e.g. by changing classes between epochs). Transfer learning is also a valid
approach for FSL. It consists in training a model on a large dataset and then
adapt it to perform well on another smaller one. This requires a supplementary
training step and is often subject to catastrophic forgetting
\cite{kirkpatrick2017overcoming}, and overfitting. It needs advanced tricks to
prevent these undesirable effects.

This performs relatively well for classification but the more challenging
detection task still lacks few-shot alternatives. Though, recent work focused on
Few-Shot Object Detection (FSOD) applying ideas from FSL literature to object
detection. The first approaches were mostly oriented toward transfer learning
and fine-tuning \cite{chen2018lstd, wang2020frustratingly} disregarding the
others FSL practices. Some other work took inspiration from meta-learning
\cite{yan2019meta} and representation learning \cite{karlinsky2019repmet}. This
is mostly applied to natural images, yet applications on remote sensing images
are scarce \cite{deng2020few, xiao2020few}. Even if object detection is hard,
applying it on remote sensing images is even harder: object's sizes can vary
greatly plus they can be arbitrarily oriented and densely packed. These
supplementary difficulties might explain why this specific topic remains mainly
untouched.

This work introduces a new few-shot learning method for object detection and
evaluates its performance on aerial images. It detects objects from only a few
examples of a class and without any fine-tuning. The main idea is inspired from
prototypical networks \cite{snell2017prototypical} which learns an embedding
function that maps images into a representation space. Prototypes are computed
from the few examples available for each class and classification scores are
attributed to each input image according to the distances between its embedding
and the prototypes. The classic Faster R-CNN framework is modified to perform
few-shot detection based on this idea. Both classification branch in Region
Proposal Network (RPN) and in Fast R-CNN are replaced by prototypical networks
to allow fast online adaptation. In addition, a few improvements are introduced
on the prototypical baseline in order to fix its weaknesses.

This paper begins with a brief overview of literature on object detection,
few-shot learning and the intersection of these two topics. Then the
prototypical Faster R-CNN architecture is presented in detail alongside with
several improvements on our baseline. Next,  the potential of the proposed
modifications throughout a series of experiences is demonstrated. Finally, the
proposed  approach is discussed with a critical eye, and it is asked whether
representation learning is suitable for object detection.

\section{Related work}
\subsection{Object detection}
During the last decade, deep learning and especially CNN have made impressive
progresses in most computer vision tasks. Object detection is no exception and
plenty of CNN based methods were proposed to address this problem. Among the
most common, YOLO \cite{redmon2016you} is a one-stage method with a trade-off on
speed. Faster R-CNN \cite{ren2015faster} is another well-known object detection
technique that focuses more on accuracy with two separate stages. Most
subsequent methods are more or less inspired from these. A common feature
between these techniques is the generation of anchor boxes. These are bounding
boxes chosen with different aspect ratios and sizes, uniformly distributed on
the image, that the networks take as reference for the regression. Recently,
some works tend to reduce the accuracy gap between one and two stages methods.
Especially, FCOS \cite{tian2019fcos} drops completely the anchors to build a
fully convolutional one stage object detection network that matches the
performance of Faster R-CNN.


As our work is mainly based on Faster R-CNN, we will describe its functioning in
details. As mentioned above, it is made of two stages: a Region Proposal Network
(RPN) and a prediction head described in \cite{girshick2015fast}. A third
component must also be mentioned, the backbone. This is a large CNN that
extracts features from images. It is often chosen as a ResNet \cite{he2016deep}
with a Feature Pyramidal Network (FPN) \cite{lin2017feature} on top. This
extracts features from different levels and is very helpful for detecting
objects of various sizes. Once the features are extracted, they are fed into the
RPN. This network is fully convolutional and will output an objectness score $
  o_a$ for each generated box $a$ (i.e. generally 3 boxes per positions in the
multi-level feature map). This score represents the likeliness of having an
object within the corresponding patch in the image. In addition, the RPN outputs
box regressions $b_a^{\text{R}}$. The regressions, combined with the anchors
sizes and positions, give the actual boxes coordinates in the image. Then, the
best scoring boxes are selected to be fed to the prediction head, where boxes
are refined and classified. For each box, the corresponding features are
selected through a pooling operation (RoI Align, proposed by \cite{he2017mask}).
These are flattened and passed to the second stage. It computes refining
coordinates shifts $b_j^{\text{H}}$ and classes scores $c_j$ for each box.
Following this, a post-processing step filters out small, low-scoring and
redundant boxes.

The training of this method is straightforward, each network has two losses, one
for the regression branch and one for the classification as described below:
\begin{align}
   & \mathcal{L}_{reg}^{\text{R}}(b_i^{\text{R}}, \hat{b}_i^{\text{R}}) &  & = \text{SmoothL1Loss}(b_i^{\text{R}}, \hat{b}_i^{\text{R}}), \\
   & \mathcal{L}_{obj}^{\text{R}}(o_i, \hat{o}_i)                       &  & = \hat{o}_i \log(o_i) + (1-\hat{o}_i) \log (1-o_i),          \\
   & \mathcal{L}_{reg}^{\text{H}}(b_j^{\text{H}}, \hat{b}_j^{\text{H}}) &  & = \text{SmoothL1Loss}(b_j^{\text{H}}, \hat{b}_j^{\text{H}}), \\
   & \mathcal{L}_{cls}^{\text{H}}(c_j, \hat{c}_j)                       &  & = - \log(c_j),
\end{align}

where SmoothL1loss is a slight modification to L1 error function. Below a
certain threshold, the error is computed according to L2 loss, while above that
threshold L1 is used. This penalizes the network less for small errors in boxes
regressions. The variables with hat correspond to ground truth values. These
losses are combined in an overall objective function that is optimized using
stochastic gradient descent:
\begin{align}
  \mathcal{L} = \mathcal{L}_{reg}^{\text{R}} + \mathcal{L}_{obj}^{\text{R}} + \mathcal{L}_{reg}^{\text{H}} + \mathcal{L}_{cls}^{\text{H}}.
\end{align}

During training, not all boxes are selected for computing losses. First, the
generated boxes are separated into two groups: positive examples, i.e. boxes
with an overlap of at least 0.7 with a ground truth annotation, and negative
examples which represent the background class. Positive and negatives examples
are included in classification losses computation while only positives are for
regression.

\subsection{Few-shot learning}
Few-shot learning corresponds to learning a task in a limited data setting.
Specifically, a task is defined as $K$-shots, $N$-ways learning when the
training set only contains $K$ examples for each of its $N$ classes. In FSL
literature, it is common to introduce the query and support sets for a given
task. The support contains the available examples: $K$ images for each of the
$N$ classes of the task. It can be seen as the training set for that task. The
query set contains images from the same classes as the support and is used for
assessing the performance of the model, like a test set.

There exists different techniques to tackle low data regime. Transfer learning
is one of them. It consists in training a network on a large-scale  dataset
(source domain) and then fine-tuning it on the few examples (target domain)
available for the actual task. This kind of methods require re-training each
time a new class is added. In the case of aerial surveillance, this is not
suitable as the adaptability must be almost immediate. In addition, these
techniques are often inclined to catastrophic forgetting
\cite{kirkpatrick2017overcoming}, a recurrent problem in continual learning.
Classes previously learned are forgotten when learning to classify new classes.
Therefore, we will only discuss in this paper methods based on meta learning or
representation learning.

Meta-learning, in the context of FSL, refers to methods that train a model
called the meta-learner whose task will be to help another model to train
efficiently (in terms of computation and data) for the actual problem. This can
be pictured as training a teacher model whose goal is to make its student learn
better. That way, the student can quickly learn new tasks and perform well in
low-data regime. Meta-learning methods often share the same training strategy
with two nested optimization loops. The inner one updates the weights of the
student net, while the outer one deals with the meta-learner. Between each
iteration of the outer loop, a new task is chosen for the inner network to
learn. The student is trained on the support set while the teacher's objective
is to maximize the student performance. Many ways of teaching were
introduced. For instance, the teacher network can be trained to directly output
weight updates of the student as described in \cite{ravi2016optimization}.
Another approach is to output only initial weights for the student as in
\cite{finn2017model}. The starting point is hopefully better than random, thus
allowing faster and better optimization. While these techniques are promising,
they do not scale very well for large networks as the teacher must be
substantially larger than the learner. In order to assess generalization
performance, classes are split into two categories: base and novel classes (or
equivalently seen and unseen classes). During meta-training, tasks are sampled
using only base classes. At test time, however they can be sampled either from
novel classes or both base and novel.

Another drawback is the necessary fine-tuning. Even if it should be quicker than
regular transfer learning, most meta-learning methods still require this
additional step. Some exceptions are based on representation learning or metric
learning. The principle is to train a network to output an abstract
representation from an input image. This is first introduced for signature
verification with siamese networks in \cite{bromley1993signature}. Prototypical
networks \cite{snell2017prototypical} is a pioneer work in using this for FSL.
An embedding network is trained to produce such representations. Before
inference, prototypes for each class are computed using the embedding net and
the support images. During inference, the query images are embedded and the
distance between their representations and the prototypes determine the
classification scores. In \cite{snell2017prototypical}, this is done with a
linear classifier, but other choices are possible. Relation networks
\cite{sung2018learning} proposes another network to compute the class scores
from the image representation and the prototypes. Similarly, matching networks
\cite{vinyals2016matching} used two separate networks to embed the image and the
prototypes. These methods are usually trained by randomly sampling tasks at each
epoch, just as other meta-learning methods. This succession of new tasks helps
the network to generalize well and therefore improves its accuracy on unseen
classes.

\begin{figure*}
  \includegraphics[width=\textwidth]{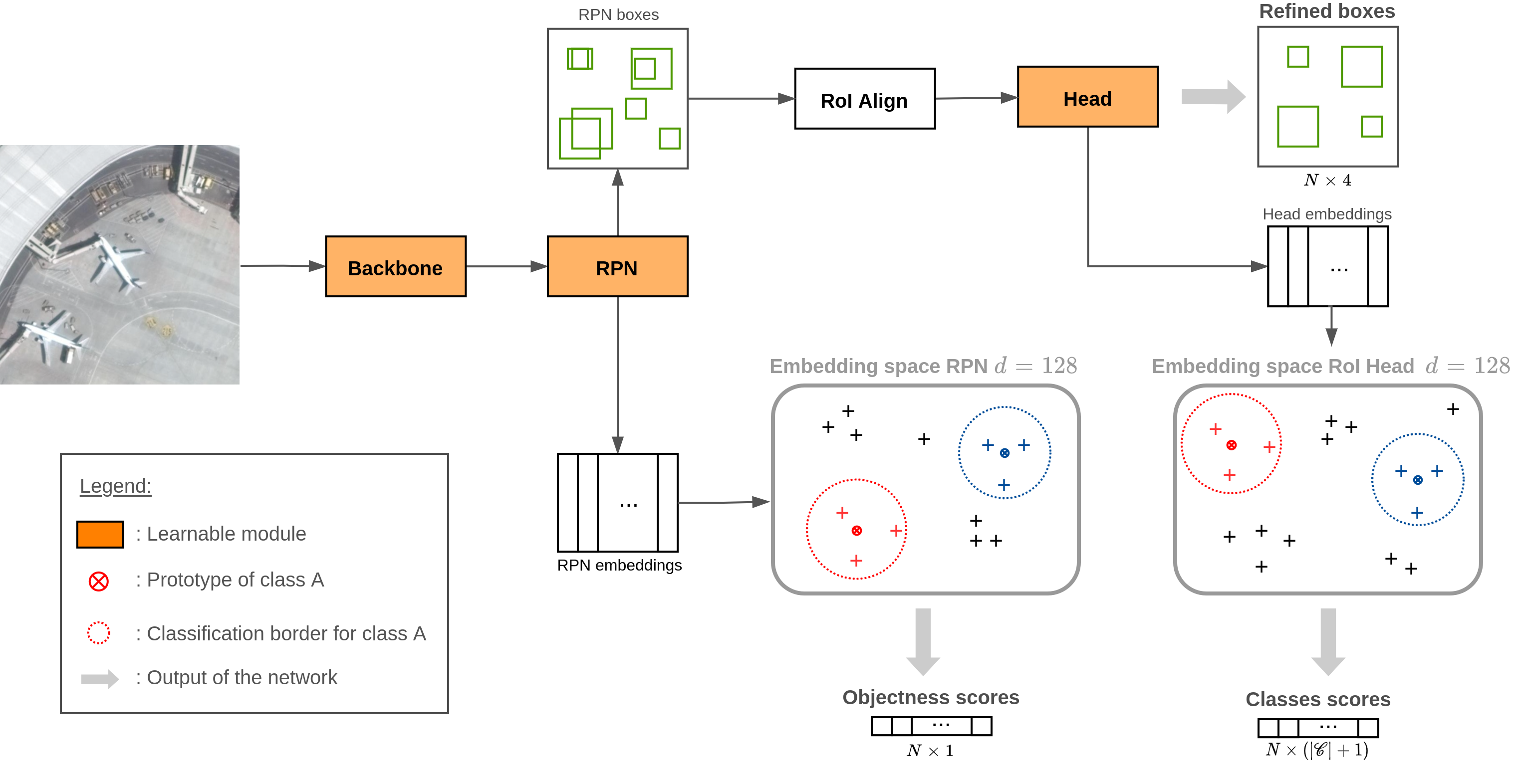}
  \caption{Architectural view of prototypical Faster R-CNN. Embedding vectors for each of the possible box locations is generated by the RPN. These are compared with class prototypes to output objectness scores. The same process occurs in the second stage to produce classes scores.}
  \label{fig:archi}
\end{figure*}

\subsection{Few shot object detection}
The focus of FSL was previously on classification tasks. Detection is a harder
problem and so is FSL. That explains why the combination of both was only
studied recently. One early work on this problem is Low-Shot Transfer Detector
\cite{chen2018lstd}. It applies few-shot transfer learning in order to refine a
pre-trained detection network on a small dataset. To do so, they introduce two
regularization losses during the fine-tuning phase. One forces the network to
focus on new objects by reducing the magnitude of the feature maps on background
areas. The other loss prevents forgetting knowledge from the source domain. It
penalizes the refined network to have dissimilar activation in the pre-softmax
layer compared to the network trained on base classes only. Similarly,
\cite{wang2020frustratingly} proposes to first pre-train a Faster R-CNN on a
base dataset and then fine-tune only the last classification and regression
layers with the new classes. This is quite similar to \cite{chen2018lstd}, but
even simpler as it does not need any new loss functions.

Even if RepMet \cite{karlinsky2019repmet} focuses mostly on few-shot
classification, authors showed that their method can also be applied for
detection. Their approach is mainly based on metric learning. During base
training, they learn alongside the network's weights a set of representatives
for different classes (several per class). These high-dimensional vectors define
the centers of Gaussian distributions inside the embedding space. The class
probabilities are assigned from the distances between the embedding vectors and
the representatives in an analogous way as in \cite{snell2017prototypical}. At
test time, the encoding network generates representatives from the few examples
in dataset, which are then fine-tuned for a few iterations. For detection, they
replace the second stage of Faster R-CNN with their classification module.
Although this method is astride on meta-learning and representation learning it
does not leverage an episodic task training strategy. Instead, it relies on
fine-tuning: it trains first on base classes and fine-tune on novel ones. To our
knowledge, this is the first attempt to solve FSOD with metric learning.

Most recent work focuses on meta-learning in order to solve FSOD. For instance,
\cite{kang2019few} trains a one stage detector along with a meta features
extractor. This extractor is a CNN that computes a reweighting vector for each
class from the support set. When a query image is passed through the detector,
the features maps output by the backbone are channel-wise multiplied with the
reweighting vectors to produce class-specific maps. These maps are then passed
to the detector's head. Each map is responsible for detecting objects of the
corresponding class. Closely related, \cite{yan2019meta} incorporates a similar
reweighting scheme in Faster R-CNN. A major difference though is that the
features and reweighting vectors are computed with the same network. A similar
class-attention mechanism, but inside the RPN network is proposed by
\cite{fan2020fsod}. In the second stage they use multi-relation heads that
combine support set information and query image features in multiple ways inside
the classification branch. Another attempt to leverage an attention mechanism is
proposed by \cite{wallach2019one}. Query and support examples are co-adapted to
reduce features discrepancies and improve prediction both in RPN and second
stage. Except the latter, these methods are trained episodically.

Different methods for combining support information with the query features
exist, e.g. \cite{xiao2020few} creates a graph between the support vectors and
the query embeddings. Then a GRU \cite{chung2014empirical} processes it in order
to provide attention between regions of interest and support set images. Alike
this, \cite{kim2020few} processes a graph, whose final nodes are the reweighting
vectors, with a graph convolution network. The graph is initialized with
similarities between classes' names, embedded with GloVe
\cite{pennington2014glove}.

With regard to FSOD on remote sensing images, very few methods exist. To the
best of our knowledge, only two works have been published on this topic.
\cite{deng2020few} improves on \cite{kang2019few} by using an improved version
of YOLO. This mainly adds multi-scale features and predictions. This is
especially important to tackle RSI as object size can vary greatly. The second
one \cite{xiao2020few}, makes use of a two-stages detector and compute
reweighting vectors with a GRU relation module. Both of the methods provide
benchmark on VHR-10 dataset \cite{cheng2016vhr}, yet the number of novel classes
and the different base/novel classes splits are different, making the comparison
difficult.

\section{Proposed method}
To deal with FSOD, prototypical Faster R-CNN is introduced. It is a modified
version of Faster R-CNN based on metric learning. The key idea is to replace the
classification branches in both stages by prototypical networks. This is related
to RepMet \cite{karlinsky2019repmet} that learns representatives only in the
second stage. Yet it fixes one major flaw of RepMet, the RPN. Once trained, the
RPN specializes on classes seen during training. This means that objects from
new classes are filtered out by the RPN, leaving no chance for the second stage
to detect them. It has a low recall on unseen classes. This is especially not
desirable in FSL. Instead, the RPN should be able to adapt to new classes. Our
method is an attempt to fill this shortcoming.

First, a few notations are introduced. Let $\mathcal{C} = [1, N]$ be the
set of all classes. In the case of $n$-ways $k$-shots learning, each task
consists in detecting objects among $n$ classes with $k$ examples per class. At
each episode a new detection task $i$ is randomly sampled: $C_i \subset
\mathcal{C}$ with $|C_i| = n$. Then, data are sampled from the whole dataset to
form a support and a query sets. $S_i = \{(x_1, t_1), ... ,(x_{nk}, t_{nk})\}$
and $Q_i = \{(x_1, t_1), ... ,(x_{nk'}, t_{nk'})\}$. For detection, each image
$x_j$ comes with a set of annotations $t_j$ which contains the location and
label of all objects in the image. Annotations that do not belong to the episode
classes are discarded. To build the support set, for each class $c \in C_i$, we
select images containing objects of class $c$ and disregard all other objects
(i.e. their annotations are not included in the support set but the image is not
masked, so they are still visible). If there are more than one object $c$ in the
image, only one is selected randomly as the annotated example. 

\subsection{Prototypical Faster R-CNN}
\label{sec:proto_faster}
We propose to change the output dimension of the classification branches in both
the RPN and the head. That way, instead of producing a classification (or
objectness) score per box, these networks output embedding vectors. Each vector
represents the information contained inside the corresponding box. The
computation of the representations is straightforward. Features at several
scales are extracted by the backbone (denoted $f$). These features are then fed
to the RPN that computes representation vectors for each location in the
features maps (i.e. each corresponding to an anchor). A 2-layers CNN (shared
with the regression branch) is applied on each feature maps, then a RoI Align
operation extracts same-size features for each anchor. Finally, a 2-layers MLP
maps these features into the embedding space. The dimension of the
representation is $r=128$ and is kept fixed in all our experiments. We call the
RPN embedding pipeline $g$ such that $(f \circ g)(x_j) = z_j$ is the set of all
anchors' embeddings for image $x_j$.

For the second stage, the best scoring boxes produced by the RPN are selected.
Their corresponding features are cropped with RoI Align and fed to a 4-layers
MLP that outputs embedding vectors as well. Note that the 2 first layers of the
head's MLP are shared with the regression branch. As for the RPN, let $h$ be the
encoding function of the head: $(f \circ h)(x_j) = \tilde z_j$. Fig.
\ref{fig:archi} illustrates the whole representation pipeline and how the
different scores are computed in the network.

Classes prototypes are computed from the support images, with the same network,
except that only the example's box is used for feature pooling. When multiple
examples are available for a class, their embeddings are averaged to build one
prototype per class: $p_{i,c}$ and $\tilde p_{i,c}$ for the RPN and the head
respectively. 

Given the embeddings of an image and the prototypes, we compute the likelihood
for each class as follows. This supposes that each class is represented by a
Gaussian distribution centered on its prototype: 
\begin{equation}
    p(x_{j,a}|y_{j,a}=c) = \exp\Big(\frac{-d(z_{j,a}, p_c)^2}{2\sigma^2}\Big),
\end{equation}
where $x_{j,a}$ refers to the crop of image $x_j$ with anchor $a$ and $y_{j,a}$
is its corresponding label ($y_{j,a} \in C_i \cup \{\varnothing\}$, with
$\varnothing$ representing the background class). $d$ is a distance measure over
the representation space, in our experiments, $d$ is the Euclidean distance.
Note that in our case, the embeddings are normalized after their computation,
therefore Euclidean distance is equivalent to Cosine Similarity. $\sigma = 0.5$
is the variance of the distribution and is fixed. This likelihood computation is
the same for the RPN and the head. However, in the head,  the likelihood of the
background class is also computed:
\begin{equation}
    p(x_{j,a} | \varnothing) = 1 - \max\limits_{c \in C_i} p(x_{j,a}| c).
\end{equation}
From this, we derive the objectness score in the RPN and the classification
(including background) scores in the head:
\begin{align}
    o_{j,a} &= \max\limits_{c \in C_i} p(x_{j,a}| c), \\
    p(c |x_{j,a}) &=\frac{p(x_{j,a}|c)}{\sum\limits_{c \in C_i \cup \{\varnothing\}} p(x_{j,a}|c)}.
\end{align}
The training is done episodically, sampling a random subset of classes $C_i
\subset \mathcal{C}$ at each epoch. The embedding network is trained using the
same loss functions as Faster R-CNN (see (1)--(5)) and the same
positive/negative examples selection. The loss is computed on the query set and
between each update of the network, the prototypes are recomputed from the same
support set. Once all query images have been seen by the network, a new task is
sampled. In our experiments, the query set contains 5 images for each of its $n$
classes, this means at least 5 examples for each class, but this number can be
larger as more than one object are present in the images. The optimization is
done with Adam optimizer and a learning rate of $1e-4$. The backbone network is
pretrained on ImageNet and its first layers are kept frozen during training.

\begin{table*}[htbp]
\centering
\caption{Mean average precision over 5 runs on DOTA dataset with 95\% confidence
interval. Results are given for two different train/test classes split. Split A:
$\{0,1,4\}$, Split B:  $\{7,11,13\}$ (only test classes are given). }
\label{tab:res_dota}
\begin{tabular}{c|c|cccc}
 & $k$ shots & 1 & 3 & 5 & 10 \\ \hline
\multirow{2}{*}{Split A} & Train classes & 0.275 $\pm$ 0.01 & 0.352 $\pm$ 0.02 & 0.390 $\pm$ 0.01 & 0.384 $\pm$ 0.02 \\
 & Test classes & 0.047 $\pm$ 0.02 & 0.024 $\pm$ 0.01 & 0.038 $\pm$ 0.01 & 0.041 $\pm$ 0.01 \\ \hline
\multirow{2}{*}{Split B} & Train classes & 0.415 $\pm$ 0.03 & 0.392 $\pm$ 0.03 & 0.434 $\pm$ 0.02 & 0.414 $\pm$ 0.03 \\
 & Test classes & 0.08 $\pm$ 0.01 & 0.101 $\pm$ 0.02 & 0.121 $\pm$ 0.01 & 0.101 $\pm$ 0.02
\end{tabular}
\end{table*}

\subsection{Iterative improvements}
\label{sec:improvements}
In order to improve the performance of our model, a series of improvements are
described on top of the baseline presented above.

\subsubsection{Hard example mining}
One issue encountered with the baseline was the detection of all training
classes regardless of support examples. This is class memorization. Although
this improves performance for training classes, it produces lots of false
positive detections. In order to address this, we propose to sample hard
negative examples to encourage the network to detect support classes only. The
main idea is to take advantage of the annotations for classes not selected in
the current task to find hard negative examples, i.e. classes that the network
could have memorized from previous tasks. With a new task at each epoch, it is
likely that the network still produces detection for objects annotated in one of
the previous epochs if it does not rely on the support information. Therefore,
these annotations can be used to find examples that should be considered as
background for the current task only. These are different from the background
examples that do not contain any class of the dataset, which are referred as
easy negative examples. Hence, this encourages the network to detect only
objects annotated in the support set.

\subsubsection{Moving average prototypes}
Another issue with the baseline is that the prototypes can change abruptly,
either when the network is updated or when the support set changes. This makes
the training unstable.  In order to prevent such rapid modification of the
prototypes, an exponential moving average is introduced to smooth the
modification. Hence, $\bar p_c^{(t+1)} = \alpha p_c + (1 - \alpha) \bar
p_c^{(t+1)}$. $\alpha$ is set to $0.1$ in our experiments. $\bar p_c^{(t)}$ is
the averaged prototype for class $c$ at iteration $t$, while $p_c^{(t)}$ is the
prototype computed from the support set, for class $c$ at iteration $t$ (as
described in section \ref{sec:proto_faster}).

\subsubsection{Background clustering}
Lastly, the baseline shows poor separation of unseen classes embeddings. This
leads to poor performance on novel classes at test time. In order to solve this,
an  inspiration is drawn from \cite{caron2018deep}. At each iteration, they fit
a K-means on the learned representations. This gives pseudo-labels to train the
network for classification in a self-supervised manner. Similarly, we propose to
fit a K-means on the negative embeddings (i.e. representing boxes not matched by
any ground truth object). From the resulting pseudo-labels a contrastive loss
function (Triplet Loss \cite{hoffer2015deep}) is computed. The triplets are
formed with embeddings that were labeled identically by the K-means. It
encourages the network to organize the embedding space into tight and separated
clusters. This will eventually discover semantic clusters that represent objects
unseen (i.e. non-annotated) during training. 

\section{Results and experiments}
In order to assess the performance of our method, one dataset is chosen: DOTA
\cite{xia2018dota} which contains remote sensing images. It has 16 different
classes and contains around 400k annotated objects. These objects are
distributed within 2800 large-scale images and their sizes vary greatly, even
within a single class as the spatial resolutions between two images can be
significantly different.

The experimental protocol is as follows: three classes are reserved for
evaluation and two different splits were randomly selected. The network is
trained episodically with the remaining classes for 30k iterations. Training
more improves the performance on training classes, but the network starts to
overfit and performs far worse on test classes. Hence, early stopping, to
preserve generalization on new classes. The networks are evaluated on both the
base and novel classes to assess both the learning and the generalization
capabilities. For each experiment, mean average precision (mAP) is provided,
computed according to PASCAL VOC \cite{everingham2010pascal}, with different
number of shots: 1, 3, 5 and 10.

\subsection{Results on DOTA}
\label{sec:res_dota}
Table \ref{tab:res_dota} contains the mean average precision reported on DOTA
dataset with our best model, according to section \ref{sec:ablation}. The
results are reported as the mean over 5 runs with 95\% confidence intervals. We
chose two different train/test classes split randomly. From this, it can be seen
that performance on test classes is far below the train classes. This is
expected since no supervision was available during training for these objects.
It is interesting to see that the more examples (i.e. shots) are given to the
network for a class, the better it performs. This may be explained as more shots
provide a better approximation of the cluster's center. Yet this pattern is not
always observed, for instance with training classes of split B, the performance
is stable with respect to the number of shots. This could happen when a single
class is represented by two separate clusters. Having multiple representation
for one class increase the chance of sampling representation from different
clusters. Hence, the average can be completely outside both clusters.
Nevertheless, it can be seen from Fig. \ref{fig:embeddings} (which was made with
one shot only) that the prototypes are often not positioned in the center of
their cluster and averaging is often a right strategy to improve performance. 

It can also be seen that the increase of performance stagnates with the number
of shots. Important gain is reported between 1, 3 and 5 shots but no significant
improvement from 5 to 10. A relatively low number of examples is able to
approximate correctly the class prototype. Increasing this number does not
improve the positioning of the centers any more and can even be harmful as
discussed above.

 


More broadly, the performance is quite low both for base and novel classes and
this does not meet our objectives. In comparison, Faster R-CNN trained in a
supervised manner on the complete dataset achieved around 0.7 mAP. It would be
unfair to directly compare this value with the performance of our network as it
was mainly designed for adaptability. Yet the performance loss on training
classes is quite large. 
 
Unfortunately, no other method proposed benchmark on DOTA dataset for FSOD and
the very few works \cite{xiao2020few, deng2020few} in this field did not provide
their code, thus comparison was not possible.

\subsection{Ablation study}
\label{sec:ablation}
In order to validate the hypothesis formulated in \ref{sec:improvements}, an
ablation study is conducted. The results of this analysis can be found in Table
\ref{tab:ablation}. They only partially validate this hypothesis. On the one
hand, the introduction of hard examples mining and moving average prototypes
improves consistently the test mAP in the one-shot setting. On the other hand,
background clustering reduces greatly the performance on train classes, while
achieving similar results on test classes. It is still unclear why this does not
perform better, it will be investigated as part of future work. According to
this analysis, we chose to fix our architecture with hard example mining and the
moving average as it combines best train and test performance. Yet, those
conclusions must be taken carefully since the variability between different runs
is high and the performance gains are small between the experiments. 

\begin{table}[htbp]
\centering
\caption{Ablation study on improvements described in section
\ref{sec:improvements}. Each row corresponds to the addition on top of the
baseline. HEM corresponds to hard example mining, MA to moving average
prototypes and BC to background clustering. Once again results are provided for
different numbers of shots, both for train and test classes. Bold scores
correspond to highest mAP for each $k$, either for train or test classes.}
\label{tab:ablation}
\begin{tabular}{c||c|cccc}
 & $k$ shots & 1 & 3 & 5 & 10 \\ \hline 
\multirow{2}{*}{Baseline} & Train & \textbf{0.355} & \textbf{0.359} & 0.343 & 0.304 \\
 & Test & 0.021 & 0.027 & 0.038 & 0.041 \\ \cline{2-6} 
\multirow{2}{*}{+ HEM} & Train & 0.312 & 0.356 & \textbf{0.412} & 0.343 \\
 & Test & 0.04 & 0.023 & 0.033 & 0.026 \\ \cline{2-6} 
\multirow{2}{*}{+ HEM + MA} & Train & 0.265 & 0.339 & 0.37 & \textbf{0.351} \\
 & Test & \textbf{0.069} & 0.035 & 0.042 & \textbf{0.059} \\ \cline{2-6} 
\multirow{2}{*}{+ HEM + MA + BC} & Train & 0.133 & 0.145 & 0.182 & 0.148 \\
 & Test & 0.043 & \textbf{0.041} & \textbf{0.047} & 0.026 \\ \hline
\end{tabular}
\end{table}

\section{Discussions and perspective}
\label{sec:discuss}
From these results, one question arises: is representation learning a suitable
choice for object detection? Representation learning methods are competitive
with state-of-the-art for few-shot classification, but seem to be inappropriate
for few-shot detection. This may be because detection task requires
distinguishing between more closely related images. When a trained RPN classifies
two overlapping patches of an image, it may produce completely different outputs
whether it contains an object entirely or not. Such quasi-discontinuities are
unlikely to occur with a randomly initialized network. Thus, the embeddings of
two closely overlapping patches will be mapped nearby in the representation
space.  



\begin{figure}
     \centering
     \begin{subfigure}[b]{0.49\columnwidth}
         \centering
         \includegraphics[width=\columnwidth, trim= 5 5 5 5, clip]{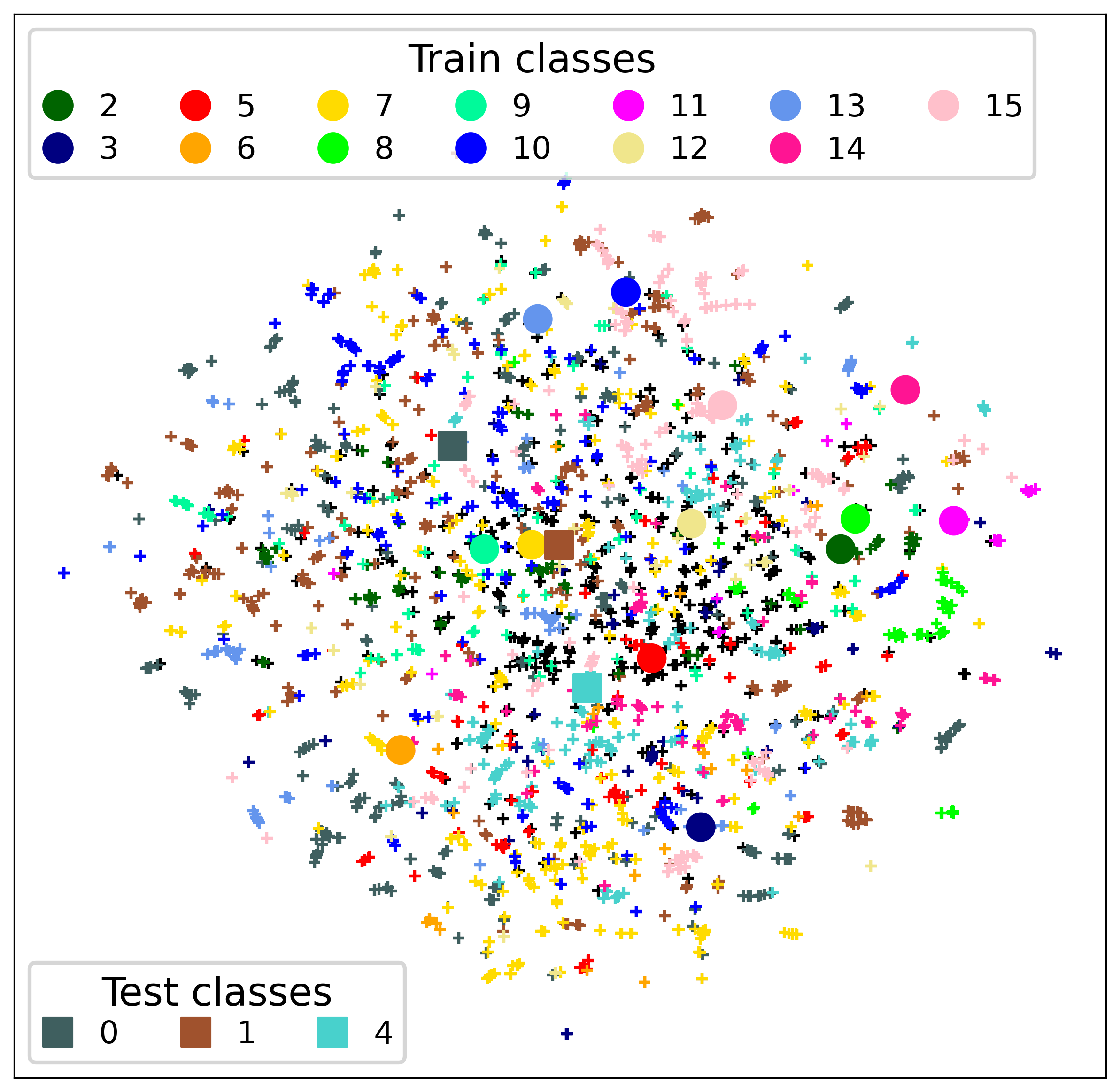}
         \caption{Before training}
         \label{fig:random_embeddings}
     \end{subfigure}
     \hfill
     \begin{subfigure}[b]{0.49\columnwidth}
         \centering
         \includegraphics[width=\columnwidth, trim= 5 5 5 5, clip]{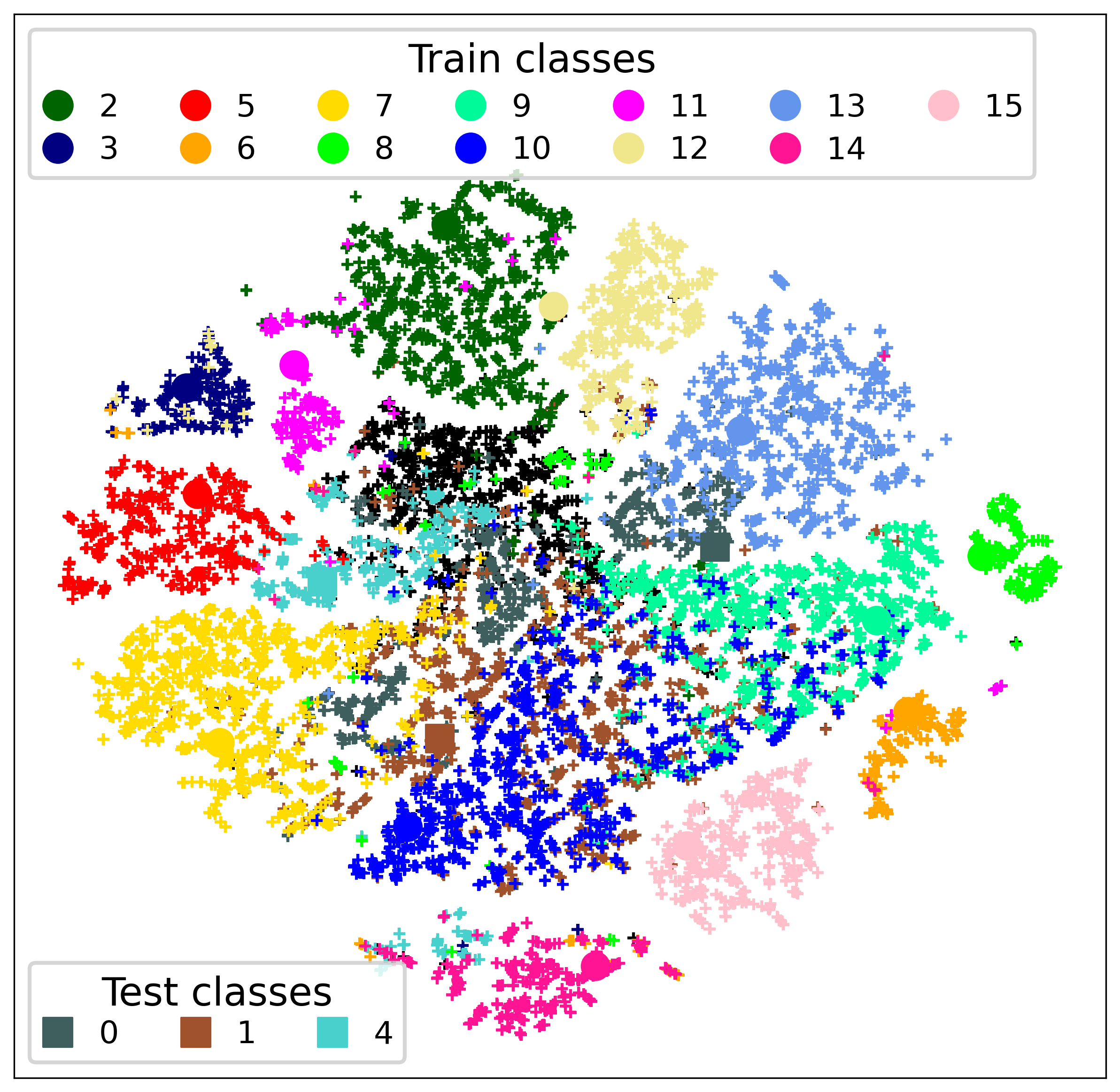}
         \caption{After training}
         \label{fig:trained_embeddings}
     \end{subfigure}
    
    \caption{TSNE visualization on the embedding space, before and after training. Training organizes 
    this space semantically and reduces the threadlike patterns representing close patches in the input
     image. Yet this is not completely solved for unseen classes during training.}
    \label{fig:embeddings}
\end{figure}

Fig. \ref{fig:random_embeddings}, where threadlike structures can be seen,
illustrates this well. These patterns are made of embeddings from close
overlapping patches within the same input image. This prior spatial organization
may be the cause of the low performance of our method. This is especially true
as the classifier on the representation space is equivalent to a linear
classifier (see \cite{snell2017prototypical} for details). For few-shot
classification, there is no such prior organizing the representation space as
two different images cannot belong to the same larger image. Hence, the
structures and colors are not as similar as those of two close patches for
object detection. Fig. \ref{fig:embeddings} shows that training is able to
overcome this and organizes the space into semantic clusters. Yet this only
happens for training classes, for which strong supervision is available during
training. For test classes, the weak supervision available is not enough to
build a semantically-aware structure: the test classes representations are mixed
together and with negative examples representations (in black in Fig.
\ref{fig:embeddings}). 

Results provided in section \ref{sec:res_dota} are computed on a test set, whose
images were not seen during training. Yet, mAP both on base and novel classes is
tracked during training on images already seen by the network. It shows a large
improvement compared to what is reported in table \ref{tab:res_dota} (around
0.65 mAP on train classes and 0.2 on test classes). This strong overfitting
showcases the lack of generalization of our method. It may be explained by a simple
reason. Objects can only represent a small part of the patch for object
detection. Hence, much information is embedded in the representation alongside
the relevant semantic information. The position of the embedding is partially
controlled by the background and the network can easily learn correlation
between background and semantic, making it easy to detect objects correctly in
previously seen images. This is especially true for small objects. In Fig.
\ref{fig:trained_embeddings}, classes 1, 9 and 10 represent small vehicles and
have the largest clusters. In comparison, classes 5, 6 and 14 respectively
represent basketball court, running track and soccer field. Their clusters are
much tighter. Their embeddings contain much variance compared to other classes
and therefore detection is harder. Of course, this could also occur for
classification but in practice the background variety is far smaller and
represent a smaller portion of the image. Hence, it produces tighter clusters
and better class separation. It can also be seen from Table \ref{tab:res_dota}
that the performance on split A (containing small vehicles) is worse than on
split B. Such a large difference between the splits suggests that the evaluation
protocol is not well-suited for this problem. More splits should be used in
order to assess generalization on all classes and performance metrics should be
reported per class.  

In order to leverage strong supervision, one could try self-supervised methods.
It has recently been shown that these methods, e.g. \cite{caron2018deep,
chen2020simple} can learn generic representations that generalize for many
visual tasks, in particular in low-data regime. It would be
interesting to investigate further these methods for few-shot object detection,
this is planned as future work. In addition, we plan to try methods based on
attention mechanism instead of representation learning as our experiments
highlight some weaknesses of the latter. Attention mechanisms, have recently
shown great performance for plenty of tasks including few-shot object detection.
Finally, a change of the underlying detection architecture is required as
modifications in Faster R-CNN can be cumbersome (due to its two stages and the
generation of anchors boxes). Instead, FCOS, which is a one-stage and
anchor-less detector, is probably better suited.

In a nutshell, we proposed a novel method for few-shot object detection based on
representation learning. These early results do not meet our expectations in
terms of performance. Yet the insights generated in this study allow to
understand the strength and weaknesses of representation learning for few-shot
object detection task. This will be helpful for future research. Ongoing work is
focusing on improving these results using simpler architecture like FCOS.



\section*{Acknowledgment}
The authors would like to thank COSE for their close collaboration and the funding of this project.
\bibliographystyle{unsrt}
\bibliography{bibliography.bib}

\end{document}